\pdfoutput=1
\documentclass[11pt]{article}

\usepackage[]{EMNLP2023}

\usepackage{times}
\usepackage{latexsym}

\usepackage[T1]{fontenc}
\usepackage[utf8]{inputenc}

\usepackage{microtype}

\usepackage{inconsolata}

\usepackage{graphicx}
\usepackage{booktabs}
\usepackage{amsthm,amsmath,amssymb,amssymb,bm,mathrsfs}
\usepackage{textcomp}
\usepackage{xcolor}
\usepackage{changes}
\usepackage{algorithm}
\usepackage{ulem}
\usepackage{subcaption}
\usepackage{multirow}
\usepackage{ulem}
\usepackage{algpseudocode}
\usepackage{float}
\usepackage{xurl}
\usepackage{enumitem}
\usepackage{amssymb}
\usepackage{cleveref}
\crefformat{section}{\S#2#1#3} 
\crefformat{subsection}{\S#2#1#3}
\crefformat{subsubsection}{\S#2#1#3}
\usepackage{arydshln}

\title{Turn-Level Active Learning for Dialogue State Tracking}

\author{Zihan Zhang\textsuperscript{1},  Meng Fang\textsuperscript{2}, Fanghua Ye\textsuperscript{3}, Ling Chen\textsuperscript{1}, Mohammad-Reza Namazi-Rad\textsuperscript{4} \\
        \textsuperscript{1}University of Technology Sydney \
        \textsuperscript{2}University of Liverpool \\
        \textsuperscript{3}University College London
        \textsuperscript{4}University of Wollongong \\
        \texttt{Zihan.Zhang-5@student.uts.edu.au}, \texttt{Meng.Fang@liverpool.ac.uk} \\
        \texttt{fanghua.ye.19@ucl.ac.uk},
        \texttt{Ling.Chen@uts.edu.au},
        \texttt{mrad@uow.edu.au} \\
        }

\begin{document}
\maketitle

\begin{abstract}

% Dialogue state tracking (DST) plays an important role in task-oriented systems. However, collecting large amount of annotated dialogue data is costly. One reason is that existing DST models treat each turn in a dialogue as an independent training instance and are trained in a fully supervised manner, thus comprehensive turn-level annotations are required. In this paper, we propose a novel \textit{turn-level} Active Learning framework for DST to iteratively select the most valuable turn in a dialogue to annotate. Experiments on MultiWOZ datasets show that our approach outperforms two strong DST baselines in the weakly-supervised scenario. Additionally, our approach can effectively achieve the same DST performance with significantly less data annotation compared to traditional training approaches.

Dialogue state tracking (DST) plays an important role in task-oriented dialogue systems. However, collecting a large amount of turn-by-turn annotated dialogue data is costly and inefficient. In this paper, we propose a novel \textit{turn-level} active learning framework for DST to actively select 
%the most valuable 
turns in dialogues to annotate. 
Given the limited labelling budget, experimental results demonstrate the effectiveness of selective annotation of dialogue turns. %show that our approach works better than other weakly-supervised scenarios, i.e., annotating the last turn, illustrating the necessity of selecting dialogue turns.
%Using different sizes of annotations, experimental results show that our approach with less annotations outperforms the standard training based on two recent DST baselines in the weakly-supervised scenarios, illustrating the efficiency of selecting dialogue turns. 
Additionally, our approach can effectively achieve comparable DST performance to traditional training approaches with significantly less annotated data, which provides a more efficient way to annotate new dialogue data\footnote{Code and data are available at 
\url{https://github.com/hyintell/AL-DST}.}.

% Dialogue state tracking (DST) plays an important role in task-oriented dialogue systems. However, collecting large amount of turn-by-turn annotated dialogue data is costly and inefficient. In this paper, we propose a novel turn-level active learning framework for DST to actively select turns in dialogues to annotate. Given the limited labelling budget, experimental results demonstrate the effectiveness of selective annotation of dialogue turns. Additionally, our approach can effectively achieve comparable DST performance to traditional training approaches with significantly less annotated data, which provides a more efficient way to annotate new dialogue data.

\end{abstract}

\section{Introduction}
\label{introduction}

% \zihan {TODO: limitations of DST: a). high cost of labelling and data scarcity, need to annotate all turns regardless the model trained using entire dialogue history or previous dialogue states b). some work try to mitigate this issue using zero-shot or few-shot learning, some propose new model that can only use last turn to train...but we are aiming to solve this issue in a more straightforward way - Active learning.. c) motivation, we believe different turn contains different level of information, such as SaCLog (difficulty), maybe give a figure of two turn examples, therefore it is natural to select important turns for training, giving limited resources to annoate each turn d) Previously, AL is typically used in classification problems, not in DST......} \\\\

Dialogue state tracking (DST) constitutes an essential component of task-oriented dialogue systems. The task of DST is to extract and keep track of the user's intentions and goals as the dialogue progresses \citep{williams-etal-2013-dialog}. Given the dialogue context, DST needs to predict all \textit{(domain-slot, value)} at each turn. Since the subsequent system action and response rely on the predicted values of specified domain-slots, an accurate prediction of the dialogue state is vital.
% \mf{talk about general things about dst}
%As shown in Figure \ref{fig_dialogue_example}, given the dialogue context, DST needs to predict all \textit{domain-slot=value} at each turn. Since the following system action and response rely on it, an accurate prediction of the dialogue state is vital.

% Despite the importance of DST, collecting annotated dialogue data for training is an expensive and time-consuming job. \citet{budzianowski-etal-2018-multiwoz} follow the Wizard-of-Oz (WoZ) \citep{10.1145/357417.357420} setup and crowd-source human workers to annotate around 10,000 multi-domain dialogues, while \citet{shah-etal-2018-bootstrapping} leverage a Machines Talking To Machines (M2M) framework to simulate user and system conversations. However, collecting large-scale dialogue datasets for data-intensive neural models is still challenging, especially for new domains.

Despite the importance of DST, collecting annotated dialogue data for training is expensive and time-consuming, and how to efficiently annotate dialogue is still challenging.
% \mf{However, how to efficiently annotate dialogue is still challenging. }
It typically requires human workers to manually annotate dialogue states \citep{budzianowski-etal-2018-multiwoz} or uses a Machines Talking To Machines (M2M) framework to simulate user and system conversations \citep{shah-etal-2018-bootstrapping}. Either way, every turn in the conversation needs to be annotated because existing DST approaches are generally trained in a fully supervised manner, where turn-level annotations are required. However, if it is possible to find the most informative and valuable turn in a dialogue to label, which enables the training of a DST model to  achieve comparable performance, we could save the need to annotate the entire dialogue, and could efficiently utilize the large-scale dialogue data collected through API calls.

% Furthermore, existing DST approaches are generally trained in a fully supervised manner, where turn-level annotations are required. They treat each turn as a single, independent training instance with no difference. However, in practice, utterances in a dialogue have different difficulty levels \citep{dai-etal-2021-preview} and do not share equal importance in a conversation.
% In addition, existing DST approaches heavily rely on fine-grained labels for training, where all domain-slots in each turn need to be annotated. However, unlabelled dialogue data is easy to obtain - in industry through API calls, but turn-level annotation requires huge efforts.

% \mf{talk about active learning instead} 

Active Learning (AL) aims to reduce annotation costs by choosing the most important samples to label \citep{settles.tr09, fang-etal-2017-learning, zhang-etal-2022-survey}. % which can be applied in this realistic scenario. 
It iteratively uses an acquisition strategy to find samples that benefit model performance the most. Thus, with fewer labelled data, it is possible to achieve the same or better performance. AL has been successfully applied to many fields in natural language processing and computer vision \citep{schumann-rehbein-2019-active, DBLP:conf/iclr/CasanovaPRP20, ein-dor-etal-2020-active, hu-neubig-2021-phrase}. However, the adoption of AL in DST has been studied very rarely. \citet{xie-etal-2018-cost} have studied to use AL to reduce the labelling cost in DST, using a  \textit{dialogue-level} strategy. They select a batch of dialogues in each AL iteration and label the entire dialogues (e.g., every turn of each dialogue), which is inefficient to scale to tremendous unlabelled data. To our knowledge, \textit{turn-level} AL remains unstudied for the task of DST. 

\begin{figure*}[!htb]
  \centering
  \includegraphics[width=\textwidth]{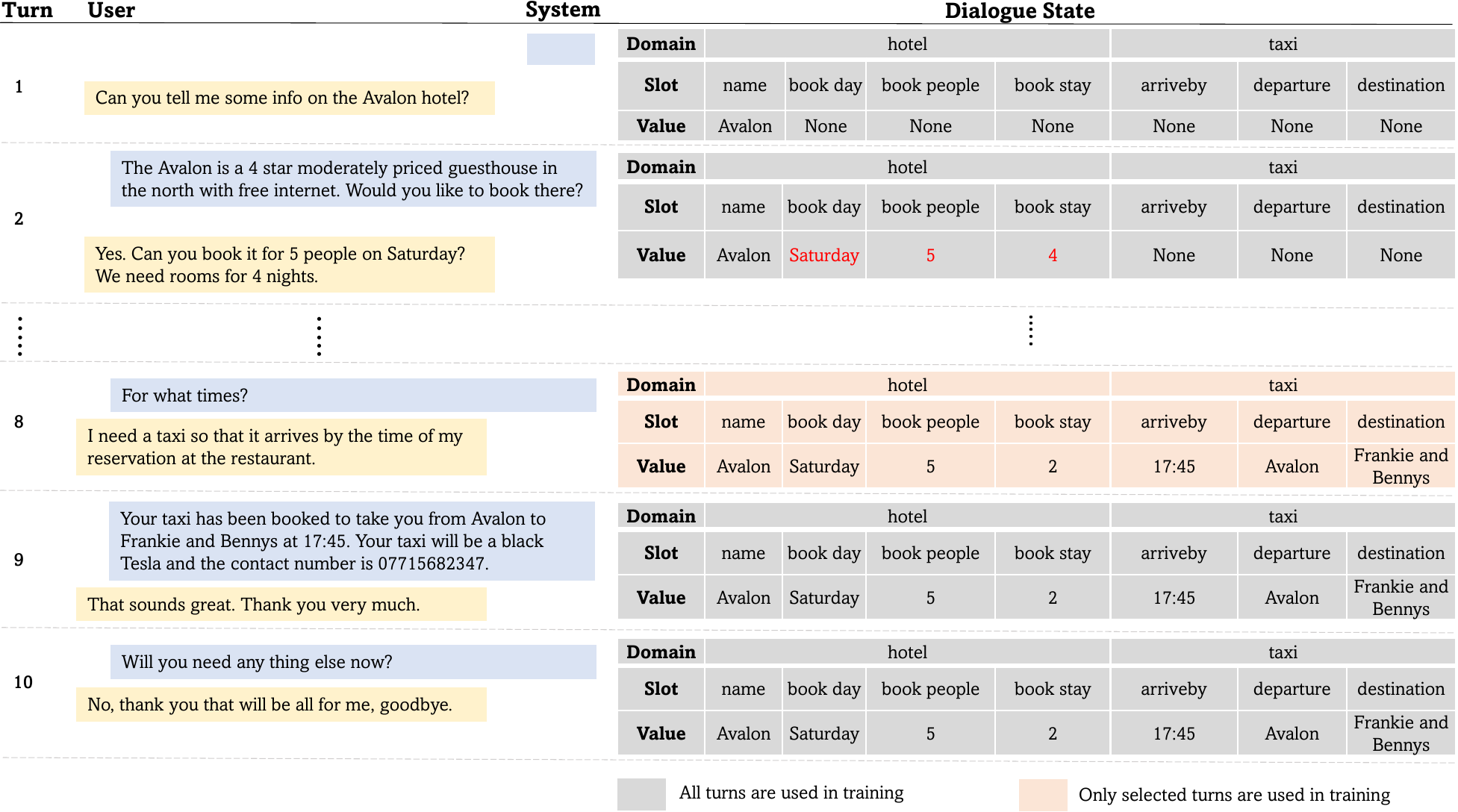}
  \caption{An example of DST from the MultiWOZ dataset \citep{budzianowski-etal-2018-multiwoz}. Utterances at the left and the right sides are from user
  and system, respectively. Orange color denotes only the selected turn is used in the weakly-supervised training setup. Only two domains (e.g \textit{hotel, taxi}) are shown due to space limitation. (best viewed in color).}
  \label{fig_dialogue_example}
\end{figure*}

Furthermore, existing DST approaches \citep{wu-etal-2019-transferable, heck-etal-2020-trippy, tian-etal-2021-amendable, zhu-etal-2022-continual}
%including  \citet{xie-etal-2018-cost},  
treat each dialogue turn as a single, independent training instance with no difference. In fact, in the real-world, utterances in a dialogue have different difficulty levels \citep{dai-etal-2021-preview} and do not share equal importance in a conversation. For example, in Fig.\ref{fig_dialogue_example}, turn-1 is simple and only contains a single domain-slot and value (i.e., \textit{hotel-name=Avalon}), while turn-2 is more complex and generates three new domain-slots, i.e., \textit{hotel-book day, hotel-book people, hotel-book stay}. Given the limited labelling budget, it is an obvious choice to label turn-2 instead of turn-1 since the former is more informative\footnote{Here, \textit{informative} refers to the turn that has more valid dialogue states.}. In addition, we observe that the complete states of the dialogue session are updated at turn-8, while turn-9 and turn-10 simply show humans' politeness and respect without introducing any new domain-slots. Therefore, while the ``last turn''  has been studied before \citep{lin-etal-2021-knowledge},  it is often not the case that only the last turn of a dialogue session generates summary states.
Using redundant turns such as turn-9 and turn-10 for training not only requires additional labelling but also possibly distracts the DST model since it introduces irrelevant context information, thus hindering the overall performance \citep{yang-etal-2021-comprehensive}. 
% \mf{any reference for the claim}

% \mf{In this paper, we consider turn-level active learning for dialogue state tracking.} 

Built on these motivations, we investigate a practical yet rarely studied problem: \textit{given a large amount of unlabelled dialogue data with a limited labelling budget, how can we annotate the raw data more efficiently and achieve comparable DST performance?} To this end, we propose a novel turn-level AL framework for DST that selects the most valuable turn from each dialogue for labelling and training. Experiments on MultiWOZ 2.0 and 2.1 show that our approach outperforms two strong DST baselines in the weakly-supervised scenarios and achieves comparable DST performance with significantly less annotated data, demonstrating both effectiveness and data efficiency. We summarize the main contributions of our work as follows:

\setlist{nolistsep}
\begin{itemize}[noitemsep]
    \item We propose a novel model-agnostic \textit{turn-level} Active Learning framework for dialogue state tracking, which provides a more efficient way to annotate new dialogue data. To our best knowledge, this is the first attempt to apply turn-level AL to DST.
    \item The superiority of our approach is twofold: firstly, our approach strategically selects the most valuable turn from each dialogue to label, which largely saves annotation costs; secondly, using significantly reduced annotation data, our method achieves the same or better DST performance under the weakly-supervised setting.
    \item We investigate how turn-level AL can boost the DST performance by analyzing the query sizes, base DST models, and turn selection strategies.
    % \item We bring the attention of AL to the DST community and show a promising research direction. \textcolor{red}{Ling: I think this one is not necessary.}
\end{itemize}

\section{Related Work}

\subsection{Dialogue State Tracking}

% % main picture
% \begin{figure*}[!htb]
%   \centering
%   \includegraphics[width=\textwidth,scale=1.5]{fig_dialogue_example.eps}
%   \caption{An example of DST from the MultiWOZ dataset \citep{budzianowski-etal-2018-multiwoz}. Utterances at the left and the right sides are from user
%   and system, respectively. Orange color denotes only the selected turn is used in the weakly-supervised training setup. Only two domains (e.g \textit{hotel, taxi}) are shown due to space limitation. (best viewed in color).}
%   \label{fig_dialogue_example}
% \end{figure*}

Dialogue state tracking is an essential yet challenging task in task-oriented dialogue systems \citep{williams-etal-2013-dialog}. Recent state-of-the-art DST models \citep{wu-etal-2019-transferable, kim-etal-2020-efficient, heck-etal-2020-trippy, 10.1145/3442381.3449939, tian-etal-2021-amendable, lee-etal-2021-dialogue,zhu-etal-2022-continual, hu-etal-2022-context} using different architectures and mechanisms have achieved promising performance on complex multi-domain datasets \citep{budzianowski-etal-2018-multiwoz, eric-etal-2020-multiwoz}. However, they are generally trained with extensive annotated data, where each dialogue turn requires comprehensive labelling. 

To mitigate the cost of dialogue annotation, some works train DST models on existing domains and perform few-shot learning to transfer prior knowledge to new domains \citep{wu-etal-2019-transferable, zhou2019multi}, while others further improve transfer learning by pre-training extensive heterogeneous dialogue corpora using constructed tasks \citep{wu-etal-2020-tod, peng-etal-2021-soloist, lin-etal-2021-zero, su-etal-2022-multi}. Recently, \citet{liang2021attention, lin-etal-2021-knowledge} propose a weakly-supervised training setup, in which only the last turn of each dialogue is used. Despite the promising results, we have shown the potential drawbacks of using the last turns in Section \ref{introduction}. In contrast, in this work, we consider the differences between the turns and strategically select the turn that benefits the DST model the most from a dialogue for training.

% To mitigate the expense of dialogue annotation, many approaches have been proposed. One trend is to train the models on some domains and perform zero-shot or few-shot learning to transfer prior knowledge to unseen domains \citep{wu-etal-2019-transferable, zhou2019multi}. However, these methods still rely on large amount of labelled data to cover enough slot categories \citep{gao-etal-2020-machine}. 

% Another line of approaches further improves transfer learning by leveraging various techniques. \citet{gao-etal-2020-machine} employ question answering (QA) data to formalize DST as a reading comprehension problem and boost few-shot performance, while \citet{lin-etal-2021-zero} further propose a unified framework to combine both extractive and multi-choice QA data to facilitate domain knowledge transfer. \citet{wu-etal-2020-tod}, \citet{peng-etal-2021-soloist}, and \citet{su-etal-2022-multi} perform task-specific pre-training on extensive heterogeneous dialogue corpora and achieve competitive performance in few-shot settings. However, they generally require large-scale external labelled dialogue data for training, which creates additional effort to collect such dialogue corpora. Recently, prompt-based methods have also shown promising results in improving few-shot learning in DST \citep{lee-etal-2021-dialogue, mi2021cins}. Despite the effectiveness of these approaches, they require additional work to construct appropriate prompting input, which is a non-trivial task.  

\subsection{Active Learning}
\label{introduction_al}

Active Learning uses an acquisition strategy to select data to minimize the labelling cost while maximizing the model performance \citep{settles.tr09}. While AL has been successfully used in many fields, such as image segmentation \citep{DBLP:conf/iclr/CasanovaPRP20}, named entity recognition \citep{shen-etal-2017-deep}, text classification \citep{schumann-rehbein-2019-active}, and machine translation \citep{zeng-etal-2019-empirical, hu-neubig-2021-phrase}, rare work has attempted to apply AL to DST. Moreover, recently proposed AL acquisition methods are,  unfortunately,  not applicable to turn-level DST since they are designed for specific tasks or models. For instance, BADGE \citep{ash2019deep} calculates gradient embeddings for each data point in the unlabelled pool and uses clustering to sample a batch, whereas we treat each turn within a dialogue as a minimum data unit and only select a single turn from each dialogue; 
% therefore the diversity-based methods do not make sense in our scenario.
therefore, the diversity-based methods are not applicable to our scenario.
ALPS \citep{yuan-etal-2020-cold} uses the masked language model loss of BERT \citep{devlin-etal-2019-bert} to measure uncertainty in the downstream text classification task, while CAL \citep{margatina-etal-2021-active} selects contrastive samples with the maximum disagreeing predictive likelihood. Both are designed for classification tasks, so these strategies are not directly applicable. Therefore, studying an AL acquisition strategy that is suitable for DST is still an open question.

% However, relatively little work has been done to incorporate AL with DST. One exception is \citet{xie-etal-2018-cost}, who is the first to use AL to reduce the labelling cost in DST. Although \citet{xie-etal-2018-cost} effectively achieve similar DST performance using significantly less annotated data, they evaluate the model on the DSTC2 \citep{henderson2014second} dataset, which is a relatively easy dataset that focuses on a single domain and has less slots. Additionally, \citet{xie-etal-2018-cost} adopt dialogue-level query strategies, i.e. they select a batch of dialogues in each iteration and label all the turns. Therefore, turn-level query strategies remain unstudied.

\section{Preliminaries}
\label{preliminaries}

We formalize the notations and terminologies used in the paper as follows.

\paragraph{Active Learning (AL)} AL aims to strategically select informative unlabelled data to annotate while maximizing a model's training performance \citep{settles.tr09}. This paper focuses on pool-based active learning, where an unlabelled data pool is available. Suppose that we have a model $\mathcal{M}$, a small seed set of labelled data $\mathcal{L}$ and a large pool of unlabelled data $\mathcal{U}$. A typical iteration of AL contains three steps: (1) train the model $\mathcal{M}$ using $\mathcal{L}$; (2) apply an acquisition function $\mathcal{A}$ to select $k$ instances from $\mathcal{U}$ and ask an oracle to annotate them; and (3) add the newly labelled data into $\mathcal{L}$.

% \paragraph{Dialogue State Tracking (DST)} Given a dialogue $D = \left\{\left(X_{1}, B_{1}\right), \ldots,\left(X_{T}, B_{T}\right)\right\}$ that contains $T$ turns, $X_t$ denotes the dialogue turn consisting of the user utterance and system response at turn $t$, while $B_t$ is the corresponding dialogue state. The dialogue state at turn $t$ is defined as $B_{t}=\left\{\left(d_j, s_{j}, v_j\right), 1 \leq j \leq J \right\}$, where $J$ denotes the total number of predefined domain-slot pairs (e.g. \textit{attraction-area}) and $v_j$ is the corresponding value of the domain-slot $(d_j, s_j)$ (e.g. \textit{south}).
% \textcolor{red}{Ling: it is not clear to me. Is $d_j$ the domain, and $s_j$ the slot? Could you provide a clearer example?}
% Given the dialogue context up to turn $t$, i.e. $H_t = \left\{\left(X_{1}, B_{1}\right), \ldots,\left(X_{t}\right)\right\}$, the objective of DST is to predict the dialogue statehttps://www.overleaf.com/project/623c6b61094f6771e6e50084 $B_t$.

\paragraph{Dialogue State Tracking (DST)} Given a dialogue $D = \left\{\left(X_{1}, B_{1}\right), \ldots,\left(X_{T}, B_{T}\right)\right\}$ that contains $T$ turns, $X_t$ denotes the dialogue turn consisting of the user utterance and system response at turn $t$, while $B_t$ is the corresponding dialogue state. The dialogue state at turn $t$ is defined as $B_{t}=\left\{\left(d_j, s_{j}, v_j\right), 1 \leq j \leq J \right\}$, where $d_j$ and $s_j$ denote domain (e.g. \textit{attraction}) and slot (e.g. \textit{area}) respectively, $v_j$ is the corresponding value (e.g. \textit{south}) of the domain-slot, and $J$ is the total number of predefined domain-slot pairs.  Given the dialogue context up to turn $t$, i.e. $H_t = \{X_{1}, \ldots,X_{t}\}$, the objective of DST is to predict the value for each domain-slot in dialogue state $B_t$. 

% Note that, by labelling the selected turn $t$,
% That is, we provide the entire dialogue utterances from the first turn to the selected turn $t$ as the input to the model, so that the information from the earlier turns is still in the dialogue history.

\paragraph{Labelling} Suppose that we have selected a turn $t$ from the dialogue $D$ ($1 \leq t \leq T$) to label. An oracle (e.g. human annotator) reads the dialogue history from $X_1$ to $X_t$ and labels the current dialogue state $B_t$.
%An oracle (e.g. human annotator) labels the current dialogue state $B_t$ given the dialogue history $\{X_1,...,X_t\}$. 
We use the gold training set to simulate a human annotator in our experiments. 

\paragraph{Full vs. Weakly-supervised Training} Generally, the training dataset for DST is built in the way that each turn in a dialogue (concatenated with all previous turns) forms  an individual training instance. That is, 
the input of a single training instance for turn $t$ is defined as $M_t = X_1 \oplus X_2 \oplus \cdots \oplus X_{t}$, where $\oplus$ denotes the concatenation of sequences, and the output is the corresponding dialogue state $B_t$. By providing the entire dialogue utterances from the first turn to turn $t$ to the model, the information from the earlier turns is kept in the dialogue history.
Let $\mathcal{D}_D$ be the set of training instances created for the dialogue $D$ and $t$ is the selected turn. Given the example in Fig.\ref{fig_dialogue_example}, for full supervision, all turns are used for training (i.e., $\mathcal{D}_D = \left\{\left(M_{1}, B_{1}\right), \ldots,\left(M_{T}, B_{T}\right)\right\}$), whereas in weakly-supervised training, only the selected turn is used (i.e., $\mathcal{D}_D = \{\left(M_{t}, B_{t}\right)\}$). 

% Note that, 
% %we only use the selected turn $t$ from $D$ when training,
% for the selected turn $t$, we provide the entire dialogue utterances from the first turn to turn $t$ as the input to the model (i.e., $\{X_1,...,X_t\}$), so that the information from the earlier turns is still in the dialogue history.

% \paragraph{Dialogue State Tracking (DST)} Given a dialogue $D = \left\{\left(U_{1}, R_{1}\right), \ldots,\left(U_{T}, R_{T}\right)\right\}$ that contains $T$ turns, $U_t$ and $R_t$ represent the user utterance and system response at turn $t$, respectively. The dialogue state at turn $t$ is defined as $B_{t}=\left\{\left(s, v_{t}\right), s \in \mathcal{S}\right\}$, where $\mathcal{S}$ denotes the set of predefined domain-slot pairs (e.g. \textit{attraction-area}) and $v_t$ is the corresponding value of the slot $s$ (e.g. \textit{south}). Given the dialogue context up to turn $t$, i.e. $H_t = \left\{\left(U_{1}, R_{1}\right), \ldots,\left(U_{t}, R_{t}\right)\right\}$, DST is required to predict the dialogue state $B_t$.

% Generally, the training dataset for DST is built in the way that each turn in a dialogue is a separate training instance, i.e. the labelled data pool $\mathcal{L} = \{D_1, D_2,..., D_N\}$, where $N$ is the total number of dialogues. Existing DST models are trained in a fully supervised manner, where every turn in a dialogue needs to be labelled by an oracle. To reduce the high cost of labelling, we apply AL to select the most valuable turn from each dialogue as a training instance to label.

\section{Active Learning for Dialogue State Tracking}

% In this section, we first define the problem setting,  followed by our turn-level AL framework and the turn selection strategies.

In this section, we first define our turn-level AL-based DST  framework, followed by the turn selection strategies.

\subsection{Turn-Level AL for DST}

%In this section, we introduce how we incorporate AL with DST, as shown in Algorithm \ref{al_algorithm}. We first train a base DST model for a fixed number of epochs if the initial labelled dialogue pool $\mathcal{L}$ is available. Otherwise, we directly use the standard initialized model in the first iteration. 

% \mf{Given a dialog $D = \left\{X_{1}, \ldots,X_{T} \right\}$, that contains $T$ turns, $X_t$ denotes the dialogue turn consisting of the user utterance and system response at turn $t$. The problem is to select a turn $X_s$ and ask for annotation $B_s$. }

\paragraph{Framework.} Our turn-level AL-based DST consists of two parts. First, we use AL to model the differences between turns in a dialogue and find the turn that is the most beneficial to label. The pseudo-code of this step is shown in Algo. \ref{al_algorithm}. Second, after acquiring all labelled turns, we train a DST model as normal and predict the dialogue states for all turns in the test set for evaluation, as described in Section \ref{preliminaries}. In this paper, we assume the training set is unlabelled and follow the cold-start setting (Algo. \ref{al_algorithm} Line 4), where the initial labelled data pool $\mathcal{L} = \emptyset$. We leave the warm-start study for future work.

% \paragraph{Active Learning Loop} In each iteration, we first randomly sample $k$ dialogues from the unlabelled pool $\mathcal{U}$, then we apply a turn acquisition function $\mathcal{A}$ and the intermediate DST model trained from the last iteration to each dialogue $D$ to select an unlabelled turn (Algo. \ref{al_algorithm} Line 10). It is noteworthy that we consider each turn within a dialogue as a minimum data unit to perform query selection. This is significantly different from \citet{xie-etal-2018-cost}, where they select \textcolor{red}{Ling: or `select'?} a few dialogues from the unlabelled pool and label all turns as the training instances.\textcolor{red}{Ling: my gut feeling here is that your turn level strategy is orthogonal to Xie et al. A brief discussion should be added to explain why not combining both strategies.} To avoid overfitting, we re-initialize the base DST model and re-train it on the current accumulated labelled data $\mathcal{L}$. After $R$ iterations, we acquire the final training set $\mathcal{L}$.

\paragraph{Active Learning Loop.} In each iteration, we first randomly sample $k$ dialogues from the unlabelled pool $\mathcal{U}$. Then, we apply a turn acquisition function $\mathcal{A}$ and the intermediate DST model trained from the last iteration to each dialogue $D$ to select an unlabelled turn (Algo. \ref{al_algorithm} Line 10). It is noteworthy that we consider each turn within a dialogue as a minimum data unit to perform query selection. This is significantly different from \citet{xie-etal-2018-cost}, where they select a few dialogues from the unlabelled pool and label all turns as the training instances. Orthogonal to \citet{xie-etal-2018-cost}'s work, it is possible to combine our turn-level strategy with dialogue-level AL. However, we leave it as future work because the AL strategies to select dialogues and turns could be different to achieve the best performance. In this work, we focus on investigating the effectiveness of AL strategies for turn selection.

To avoid overfitting, we re-initialize the base DST model and re-train it on the current accumulated labelled data $\mathcal{L}$. After $R$ iterations, we acquire the final training set $\mathcal{L}$.

% In other words, we aim to select a single turn from each dialogue in $\mathcal{U}$ that is the most valuable to label.  After $R$ iterations, we acquire $N$ training instances, which is the total number of dialogues.

\begin{algorithm}
    \caption{Turn-level AL for DST}
    \label{al_algorithm}
    \begin{algorithmic}[1] % The number tells where the line numbering should start
        \Require{Initial DST model $\mathcal{M}$, unlabelled dialogue pool $\mathcal{U}$, labelled data pool $\mathcal{L}$,  number of queried dialogues per iteration $k$, acquisition function $\mathcal{A}$, total iterations $R$}
        \If {$\mathcal{L} \neq \emptyset$}
            \State  $\mathcal{M}_0 \gets $ Train $\mathcal{M}$ on $\mathcal{L}$ \Comment{Warm-start}
        \Else 
            \State $\mathcal{M}_0 \gets \mathcal{M}$ \Comment{Cold-start}
        \EndIf
        \For {iterations $r = 1:R$} 
            \State $\mathcal{X}_r = \emptyset$
            \State $\mathcal{U}_r \gets$ Random sample $k$
            dialogues from $\mathcal{U}$
            \For {dialogue $D \in \mathcal{U}_r$}
                \State $X \gets \mathcal{A}(\mathcal{M}_{r-1}, D)$ \Comment{Select a turn}
                \State $\mathcal{X}_r = \mathcal{X}_r \cup \{X\}$
            \EndFor
            \State $\mathcal{L}_r \gets$ Oracle labels $\mathcal{X}_r$
            \State $\mathcal{L} = \mathcal{L} \cup \mathcal{L}_r$
            \State $\mathcal{U} = \mathcal{U} \setminus \mathcal{U}_r$
            \State $\mathcal{M}_r \gets $ Re-initialize and re-train $\mathcal{M}$ on $\mathcal{L}$
        \EndFor
        \State \textbf{return} $\mathcal{L}$ \Comment{The final training set}
    \end{algorithmic}
\end{algorithm}

\subsection{Turn Selection Strategies}
%\subsection{Active Turn Selection}
\label{active_turn_selection}

As mentioned in Section \ref{introduction_al}, recently proposed AL acquisition strategies are not applicable to DST. Therefore, we adapt the common uncertainty-based acquisition strategies to select a turn from a dialogue:

\paragraph{Random Sampling (RS)} We randomly select a turn from a given dialogue. Despite its simplicity, RS acts as a strong baseline in literature \citep{settles.tr09, xie-etal-2018-cost, ein-dor-etal-2020-active}.
\begin{equation}
% \small
    X = \mathrm{Random}(M_{1}, \ldots, M_{T})
    \label{eq1}
\end{equation}
where $T$ is the total number of turns in the dialogue.

% \paragraph{Last Turn (LT)} We choose the last turn of a given dialogue based on the assumption that the last turn contains the complete dialogue states and is the most informative. This is similar to the weakly supervised setting in \citep{liang2021attention, lin-etal-2021-knowledge}, but is in an iterative manner.
% \begin{equation}
%     X = X_{T} 
%     \label{eq2}
% \end{equation}

\paragraph{Maximum Entropy (ME)}\citep{lewis1994sequential} Entropy measures the prediction uncertainty of the dialogue state in a dialogue turn. %Different from \citet{xie-etal-2018-cost}, 
In particular, we calculate the entropy of each turn in the dialogue and select the highest one. To do that, we use the base DST model to predict the value of the $j$th domain-slot at turn $t$, which gives us the value prediction distribution $\mathbf{P}_{t}^{j}$. We then calculate the entropy of the predicted value using $\mathbf{P}_{t}^{j}$ (Eq.\ref{eq2}):

% the prediction distribution \textcolor{red}{Ling: what is prediction distribution? or $\mathbf{P}_{t}^{j}$?} of the $j$th domain-slot at turn $t$ is $\mathbf{P}_{t}^{j}$, and its entropy is calculated as:

{
% \small
\begin{align}
    &\mathbf{e}^j_t = -\sum_{i=1}^{V} \mathbf{p}_{t}^{j}[i] \log \mathbf{p}_{t}^{j}[i] \label{eq2} \\
    &\mathbf{e}_t = \sum_{j=1}^{J} \mathbf{e}^j_t \label{eq3} \\
    &X = \mathrm{argmax}(\mathbf{e}_1, \ldots,\mathbf{e}_T)
    \label{eq4}
\end{align}
}%
where $V$ is all possible tokens in the vocabulary. We then sum the entropy of all domain-slots as the turn-level entropy (Eq.\ref{eq3}) and select the maximum dialogue turn (Eq.\ref{eq4}).

\paragraph{Least Confidence (LC)} LC typically selects instances where the most likely label has the lowest predicted probability \citep{culotta2005reducing}. In DST, we use the sum of the prediction scores for all $J$ domain-slots to measure the model's confidence when evaluating a dialogue turn, and select the turn with the  minimum confidence:

{
% \small
\begin{align}
    % \begin{aligned}
    & \mathbf{c}_t = \sum_{j=1}^{J} \mathbf{c}^j_t \\ 
    & X = \mathrm{argmin}(\mathbf{c}_1, \ldots, \mathbf{c}_T)
    % \end{aligned}
    \label{eq6}
\end{align}
}%
where $\mathbf{c}^j_t = \mathrm{max}(\mathrm{logits}^j_t)$ denotes the maximum prediction score of the $j$th domain-slot at turn $t$ and $\mathrm{logits}^j_t$ is the predictive distribution.

% We use \textbf{KAGE-GPT2}\footnote{We use the best model setting \texttt{L4P4K2-DSGraph} as in the original paper.} \citep{lin-etal-2021-knowledge} as the base DST model to implement all experiments. KAGE-GPT2 is a generative model that incorporates a Graph Attention Network to explicitly learn the relationships between domain-slots before predicting slot values. It shows state-of-the-art performance in both full and weakly-supervised scenarios on MultiWOZ2.0 \citep{budzianowski-etal-2018-multiwoz}. 

% To show the effectiveness of incorporating AL is not tied to specific base models, we also experiment with an end-to-end task-oriented dialogue model \textbf{PPTOD}\footnote{We use the pre-trained PPTOD$_{\text{base}}$ from its official release \url{https://github.com/awslabs/pptod}.} \citep{su-etal-2022-multi}. PPTOD pre-trained on large dialogue corpora and gains competitive results on DST in the low-resource settings.

% \subsection{Algorithm}

% \subsection{Training}

% After acquiring the training set, we train the base DST model by following its original hyperparameter settings. We train the DST model for larger epochs and save the best model based on the performance on the validation set. 

\section{Experiments}

\subsection{Setup}
\paragraph{Datasets.}
% We evaluate the weakly-supervised DST performance on MultiWOZ 2.0 \citep{budzianowski-etal-2018-multiwoz} and MultiWOZ 2.1 \citep{eric-etal-2020-multiwoz}. MultiWOZ 2.0 is one of the largest multi-domain task-oriented dialogue datasets, including more than 10,000 dialogues spanning around seven domains. MultiWOZ 2.1 is a rectified version of MultiWOZ 2.0, which fixes annotation errors. We use the same preprocessing as \citet{lin-etal-2021-knowledge} and \citet{su-etal-2022-multi}, and focus on five domains (i.e. \textit{restaurant, train, hotel, taxi, attraction}). The statistics of the datasets are summarized in Table \ref{datasets_statistics}.

We evaluate the weakly-supervised DST performance on the MultiWOZ 2.0 \citep{budzianowski-etal-2018-multiwoz} and MultiWOZ 2.1 \citep{eric-etal-2020-multiwoz} datasets\footnote{We also tried to use the SGD dataset \citep{Rastogi_Zang_Sunkara_Gupta_Khaitan_2020}. However, the PPTOD model is already pre-trained on this dataset, making it unsuitable for downstream evaluation. KAGE-GPT2 requires the predefined ontology to build a graph neural network, but SGD does not provide all possible values for non-categorical slots (See Section \ref{limitations}).} as they are widely adopted in DST. 
% They are one of the largest multi-domain task-oriented dialogue datasets, including more than 10,000 dialogues spanning around seven domains. 
We use the same preprocessing as \citet{lin-etal-2021-knowledge} and \citet{su-etal-2022-multi}, and focus on five domains (i.e. \textit{restaurant, train, hotel, taxi, attraction}). The statistics of the datasets are summarized in Appendix \ref{appendix_datasets}.
\begin{table*}[ht!]
\centering
\resizebox{1.0\textwidth}{!}{
\begin{tabular}{llcccccc}
%\hline
\toprule
\multirow{2}{*}{\textbf{Training Data}} & \multirow{2}{*}{\textbf{Model}} & \multicolumn{3}{c}{\textbf{MultiWOZ 2.0}} & \multicolumn{3}{c}{\textbf{MultiWOZ 2.1}} \\ \cline{3-8} 
 &  & JGA $\uparrow$ & SA $\uparrow$ & RC $\downarrow$ & JGA $\uparrow$ & SA $\uparrow$ & RC $\downarrow$ \\ \midrule %\hline

\noalign{\vskip 1.5mm}
\multicolumn{1}{c}{} & \multicolumn{1}{c}{} & \multicolumn{6}{c}{\textbf{\textit{Without Active Learning}}} \\ \toprule  %\hline

%%%%%%%%%%%%%%%%%%%%
% full data section
%%%%%%%%%%%%%%%%%%%%
% \multirow{2}{*}{\textbf{Full Data (100\%)}} & 
%   PPTOD$_{\text{base}}$ & 53.37$^\ddagger$ & 97.26\normalsize{$\pm0.02$} & - & 57.10$^\ddagger$ & 97.94\normalsize{$\pm0.02$} & - \\
% % KAGE-GPT2
%  & KAGE-GPT2 & 54.86$^\dagger$ & 97.47$^\dagger$ & - & 52.13\normalsize{$\pm0.89$} & 97.18\normalsize{$\pm0.02$} & - \\ \midrule %\hline
\multirow{2}{*}{\textbf{Full Data (100\%)}} & 
  PPTOD$_{\text{base}}$ & 53.37\footnotesize{$\pm0.46$} & 97.26\footnotesize{$\pm0.02$} & 100 & 57.10\footnotesize{$\pm0.51$} & 97.94\footnotesize{$\pm0.02$} & 100 \\

\cdashline{2-8}

% KAGE-GPT2
 & KAGE-GPT2 & 54.86\footnotesize{$\pm0.12$} & 97.47\footnotesize{$\pm0.02$} & 100 & 52.13\footnotesize{$\pm0.89$} & 97.18\footnotesize{$\pm0.02$} & 100 \\ \midrule %\hline
%%%%%%%%%%%%%%%%%%%%
% Random Turn without AL section
%%%%%%%%%%%%%%%%%%%% 
%  \multirow{2}{*}{\textbf{Random Turn (14.4\%)}} & 
% % PPTOD_base random turn
% PPTOD$_{\text{base}}$-RandomTurn & 44.61\normalsize{$\pm2.19$} & 96.11\normalsize{$\pm0.05$} & 58.66\normalsize{$\pm28.1$} & 45.21\normalsize{$\pm1.55$} & 97.26\normalsize{$\pm0.03$} & 57.96\normalsize{$\pm28.7$} \\
% % KAGE-GPT2 random turn
%  & KAGE-GPT2-RandomTurn & 49.37\normalsize{$\pm0.47$} & 96.94\normalsize{$\pm0.01$} & 58.18\normalsize{$\pm28.8$} & 48.98\normalsize{$\pm0.48$} & 97.00\normalsize{$\pm0.01$} & 58.64\normalsize{$\pm28.6$} \\ \midrule %\hline
%%%%%%%%%%%%%%%%%%%%
% Last Turn without AL section
%%%%%%%%%%%%%%%%%%%%
\multirow{2}{*}{\textbf{Last Turn (14.4\%)}} & 
% PPTOD_base last turn
PPTOD$_{\text{base}}$-LastTurn & 43.83\footnotesize{$\pm1.55$} & 96.87\footnotesize{$\pm0.06$} & 100 & 45.94\footnotesize{$\pm0.72$} & 97.11\footnotesize{$\pm0.04$} & 100 \\
% % KAGE-GPT2 last turn
%  & KAGE-GPT2-LastTurn & 50.43$^\dagger$ & 97.14$^\dagger$ & 100 & 49.12\normalsize{$\pm0.13$} & 97.05\normalsize{$\pm0.02$} & 100 \\ \bottomrule %\hline

\cdashline{2-8}

% KAGE-GPT2 last turn
 & KAGE-GPT2-LastTurn & 50.43\footnotesize{$\pm0.23$} & 97.14\footnotesize{$\pm0.01$} & 100 & 49.12\footnotesize{$\pm0.13$} & 97.05\footnotesize{$\pm0.02$} & 100 \\ \bottomrule %\hline

\noalign{\vskip 1.5mm}    
\multicolumn{1}{c}{} & \multicolumn{1}{c}{} & \multicolumn{6}{c}{\textbf{\textit{With Active Learning}} ($k=2000$)} \\ \toprule %\hline 

%%%%%%
% Xie's
%%%%%
\multirow{2}{*}{\textbf{CUDS ($\sim$14\%)$^*$}} & 
PPTOD$_{\text{base}}$+CUDS & 43.06\footnotesize{$\pm0.04$} & 96.01\footnotesize{$\pm0.02$} & 100 & 43.57\footnotesize{$\pm1.16$} & 96.16\footnotesize{$\pm0.01$} & 100 \\

\cdashline{2-8}

% KAGE-GPT2 last turn
 & KAGE-GPT2+CUDS & 47.06\footnotesize{$\pm1.43$} & 96.14\footnotesize{$\pm0.07$} & 100 & 47.56\footnotesize{$\pm1.07$} & 96.33\footnotesize & 100 \\ \midrule %\hline
%%%%%%%%%%%%%%%%%%%%
% Selected Turn with AL section
%%%%%%%%%%%%%%%%%%%%
\multirow{6}{*}{\textbf{Selected Turn (14.4\%) (Ours)}} & 
% PPTOD_base 
PPTOD$_{\text{base}}$+RS & 43.71\footnotesize{$\pm0.81$} & 96.64\footnotesize{$\pm0.08$} & 58.73\footnotesize{$\pm28.7$} & 46.96\footnotesize{$\pm0.18$} & 96.56\footnotesize{$\pm0.06$} & \textbf{58.55}\footnotesize{$\pm28.5$} \\
& PPTOD$_{\text{base}}$+LC & 45.79\footnotesize{$\pm0.35$} & 97.06\footnotesize{$\pm0.04$} & 85.21\footnotesize{$\pm19.7$} & 47.37\footnotesize{$\pm0.32$} & 96.97\footnotesize{$\pm0.05$} & 81.95\footnotesize{$\pm24.6$} \\
& PPTOD$_{\text{base}}$+ME & \textbf{46.92}\footnotesize{$\pm0.79$} & \textbf{97.12}\footnotesize{$\pm0.05$} & \textbf{57.37}\footnotesize{$\pm32.9$} & \textbf{48.21}\footnotesize{$\pm1.00$} & \textbf{97.33}\footnotesize{$\pm0.12$} & 67.68\footnotesize{$\pm30.1$} \\

\cdashline{2-8}

& KAGE-GPT2+RS & 50.37\footnotesize{$\pm0.52$} & 97.11\footnotesize{$\pm0.06$} & \textbf{58.58}\footnotesize{$\pm28.7$} & 46.98\footnotesize{$\pm0.64$} & 96.81\footnotesize{$\pm0.07$} & \textbf{58.48}\footnotesize{$\pm28.5$} \\
& KAGE-GPT2+LC & 50.56\footnotesize{$\pm0.07$} & 97.10\footnotesize{$\pm0.01$} & 70.51\footnotesize{$\pm30.3$} & 48.13\footnotesize{$\pm0.20$} & 96.94\footnotesize{$\pm0.01$} & 79.41\footnotesize{$\pm24.0$} \\
& KAGE-GPT2+ME & $\textbf{51.34\footnotesize{$\pm0.05$}}$ & \textbf{97.16}\footnotesize{$\pm0.05$} & 62.58\footnotesize{$\pm28.5$} & \textbf{50.02}\footnotesize{$\pm1.10$} & \textbf{97.13}\footnotesize{$\pm0.10$} & 71.02\footnotesize{$\pm26.7$} \\ %\hline
\bottomrule
\end{tabular}
}
\caption{The mean and standard deviation of joint goal accuracy (\%), slot accuracy (\%) and reading cost (\%) after the final AL iteration on the test sets. 
% $\uparrow$: the higher the better, $\downarrow$: the lower the better. 
% $^\dagger$ and $^\ddagger$ results are cited from \citet{lin-etal-2021-knowledge} and \citet{su-etal-2022-multi} respectively, 
$^*$: we re-implement using \citet{xie-etal-2018-cost}'s method. \textbf{RS}, \textbf{LC} and \textbf{ME} are active turn selection methods mentioned in Section \ref{active_turn_selection}.
Note that we take JGA and RC as primary evaluation metrics since SA is indistinguishable \citep{kim-etal-2022-mismatch}.
}
\label{main_results}
\end{table*}

%\subsection{Base DST Model}
\paragraph{Base DST Model.}
We use \textbf{KAGE-GPT2} \citep{lin-etal-2021-knowledge} as the base DST model to implement all experiments. KAGE-GPT2 is a generative model that incorporates a Graph Attention Network to explicitly learn the relationships between domain-slots before predicting slot values. It shows strong performance in both full and weakly-supervised scenarios on MultiWOZ 2.0 \citep{budzianowski-etal-2018-multiwoz}. 
To show that the effectiveness of our AL framework is not tied to specific base models, we also experiment with an end-to-end task-oriented dialogue model \textbf{PPTOD} \citep{su-etal-2022-multi}. PPTOD pre-trained on large dialogue corpora gains competitive results on DST in the low-resource settings. The model training and implementation details are in Appendix \ref{appendix_config_details}.

\subsection{Evaluation Metrics}

% We use two commonly applied metrics to evaluate DST performance and propose a new metric to measure the cost of labelling a turn:
We use \textbf{Joint Goal Accuracy (JGA)} to evaluate DST performance, which is the ratio of correct dialogue turns. It is a strict metric since a turn is considered as correct if and only if all the slot values are correctly predicted. 
Following the community convention, although it is not a distinguishable metric \citep{kim-etal-2022-mismatch}, we also report \textbf{Slot Accuracy (SA)}, which compares the predicted value with the ground truth for each domain-slot at each dialogue turn.
% \textbf{Joint Goal Accuracy (JGA)} is the ratio of correct dialogue turns. It is a more strict metric since a turn is considered as correct if and only if all the slot values are correctly predicted. 
Additionally, we define a new evaluation metric, \textbf{Reading Cost (RC)}, which measures the number of turns a human annotator needs to read to label a dialogue turn. As shown in Fig.\ref{fig_dialogue_example}, to label the dialogue state $B_t$ at turn $t$, a human annotator needs to read through the dialogue conversations from $X_1$ to $X_t$ to understand all the domain-slot values that are mentioned in the dialogue history:
\begin{equation}
% \small
    \text{RC} = \frac{\sum_{i=1}^{|\mathcal{L}|}  \frac{t}{T_{D_i}}}{|\mathcal{L}|}
    % \label{eq1}
\end{equation}
where $|\mathcal{L}|$ denotes the total number of annotated dialogues and $T_{D_i}$ is the number of turns of the dialogue $D_i$. If all last turns are selected, then $\text{RC}=1$, in which case the annotator reads all turns in all dialogues to label, resulting high cost. 
Note that we take JGA and RC as primary evaluation metrics.

% $\text{RC}=1$ means the last turn is selected, in which case the annotator reads all turns in all dialogues. \textcolor{red}{Ling: in which case the annotator reads all turns in all dialogues}.

\subsection{Baselines}

Our main goal is to use AL to actively select the most valuable turn from each dialogue for training, therefore reducing the cost of labelling the entire dialogues. %To show the effectiveness of our approach, we compare DST performance of three settings \textit{without} involving AL:
We evaluate the effectiveness of our approach from two angles. First, we compare DST performance of two settings \textit{without} involving AL to show the benefits that AL brings: 

\begin{itemize}
    \item \textbf{Full Data (100\%)}: all the turns are used for training, which shows the upper limit of the base DST model performance.
    \item \textbf{Last Turn (14.4\%\footnote{14.4\% = $\frac{\text{\# turns used}}{\text{\# total turns}} = \frac{7888}{54945}$})}: following \citet{liang2021attention} and \citet{lin-etal-2021-knowledge}, for each dialogue, only the last turn is used for training. 
    % This acts as the lower limit \textcolor{red}{Ling: why this is the lower limit} of the DST model as the complete context information is fed into the model.
    % \item \textbf{Random Turn (14.4\%)}: for each dialogue, we randomly select a turn for training. 
\end{itemize}
%and DST performance with turns strategically selected by our turn-level AL framework:
Second, when using AL, we compare our turn-level framework with the dialogue-level approach:
\begin{itemize}
    \item \textbf{CUDS ($\sim$14\%)} \citep{xie-etal-2018-cost}: a dialogue-level method that selects a batch of dialogues in each AL iteration based on the combination of labelling cost, uncertainty, and diversity, and uses all the turns for training. We carefully maintain the number of selected dialogues in each iteration so that the total number of training instances is roughly the same (i.e., $k \simeq 2000$) for a fair comparison. 
    \item \textbf{Selected Turn (14.4\%)}: we apply Algo.\ref{al_algorithm} and set $\mathcal{U} = 7888$, $\mathcal{L} = \emptyset$, $k = 2000$ and use the turn selection methods mentioned in Section \ref{active_turn_selection} to conduct experiments. As a trade-off between computation time and DST performance, here we use $k=2000$; however, we find that a smaller $k$ tends to have a better performance (Section \ref{ablation_study}).
    % as the ablative study shown in Section \ref{ablation_study}.
    %conduct an ablative study on the size of $k$ in Section \ref{ablation_query_size}. 
    Given $k=2000$, we have selected 7,888 turns after four rounds, and use them to train a final model.
\end{itemize}

%%%%%%%%%%%%% implementation %%%%%%%%%%%%%

% \subsection{Implementation Details}

% % We use KAGE-GPT2\footnote{\url{https://github.com/LinWeizheDragon/Knowledge-Aware-Graph-Enhanced-GPT-2-for-Dialogue-State-Tracking}} and PPTOD$_{\text{base}}$\footnote{\url{https://github.com/awslabs/pptod}} from their publicly released implementations and follow their hyperparameter settings. For KAGE-GPT2, we use the best model setting \texttt{L4P4K2-DSGraph} as in the original paper; for PPTOD, we use its \texttt{base} pre-trained checkpoint. More details are in Appendix \ref{appendix_config_details}.

% We use KAGE-GPT2\footnote{\url{https://github.com/LinWeizheDragon/Knowledge-Aware-Graph-Enhanced-GPT-2-for-Dialogue-State-Tracking}} and PPTOD$_{\text{base}}$\footnote{\url{https://github.com/awslabs/pptod}} from their publicly released implementations and follow their hyperparameter settings. More details are in Appendix \ref{appendix_config_details}.

% In each AL iteration, we train a re-initialized\footnote{For PPTOD$_{\text{base}}$, we re-initialize it from the pre-trained checkpoint.} DST model for 150 epochs using the current accumulated labelled pool $\mathcal{L}$, and early stop when the performance is not improved for 5 epochs on the validation set. Importantly, instead of using the full 7,374 validation set, we only use the last turn of each dialogue to simulate the real-world scenario, where a large amount of annotated validation set is also difficult to obtain \citep{perez2021true}. However, we use the full test set when evaluating. For all experiments, we run with three different random seeds and report the average results.

\section{Results \& Analysis}

\subsection{Main Results}
\label{sec_main_results}

\begin{figure*}
  \centering
  \includegraphics[scale=0.6]{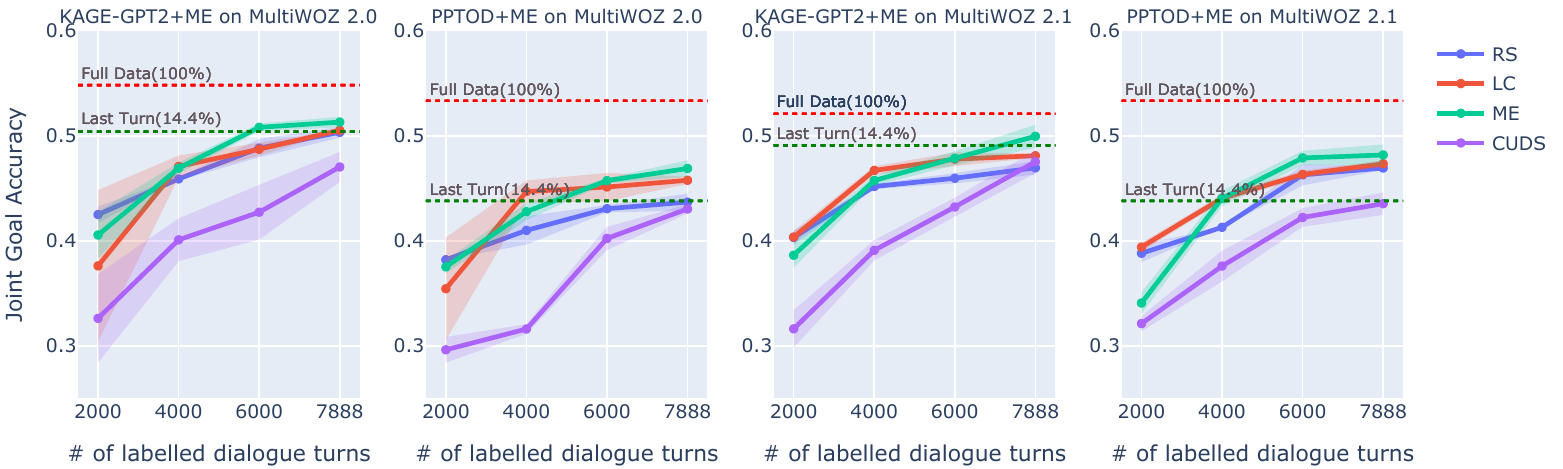}
  \caption{Joint goal accuracy on test sets of AL over four iterations with $k=2000$ dialogues queried per iteration.}
  \label{k2000_4plot}
\end{figure*}

\begin{figure}[hbt]
  \centering
  \includegraphics[width=0.9\columnwidth]{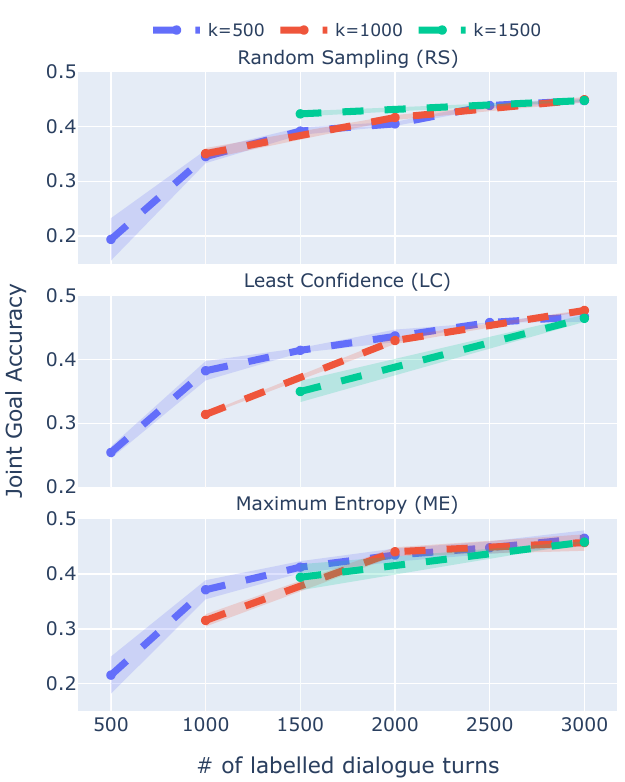}
  \caption{Joint goal accuracy on test sets of KAGE-GPT2 on MultiWOZ 2.0 with $k=500, 1000, 1500$.}
  \label{ablation_k_size}
\end{figure}

Due to space limitation, we report the final results after the four AL iterations in Table \ref{main_results}. We present the intermediate results in Fig.\ref{k2000_4plot}.

% From Table \ref{main_results}, we first observe that, using the same number of training data (14.4\%), our proposed AL approach (i.e. \texttt{PPTOD$_{\text{base}}$+ME} and \texttt{KAGE-GPT2+ME}) outperforms the two non-AL settings, \textbf{Random Turn} and \textbf{Last Turn}, in terms of both joint goal accuracy and slot accuracy. Specifically, compared with \texttt{PPTOD$_{\text{base}}$+RandomTurn} and \texttt{PPTOD$_{\text{base}}$+LastTurn}, our \texttt{PPTOD$_{\text{base}}$+ME} significantly boosts the joint goal accuracy by 2.3\% / 3.1\% on MultiWOZ 2.0, and 3.0\% / 2.3\% on MultiWOZ 2.1. \texttt{KAGE-GPT2+ME} also improves its baselines by around 0.9\% on both datasets. Compared with the dialogue-level AL strategy \textbf{CUDS}, our turn-level methods improve the JGA by a large margin.

\paragraph{Our \textit{turn-level} AL strategy improves DST performance.} From Table \ref{main_results}, we first observe that, using the same amount of training data (14.4\%), our proposed AL approach (i.e. \texttt{PPTOD$_{\text{base}}$+ME} and \texttt{KAGE-GPT2+ME}) outperforms the non-AL settings, \textbf{Last Turn}, in terms of both joint goal accuracy and slot accuracy. Specifically, compared with \texttt{PPTOD$_{\text{base}}$+LastTurn}, our \texttt{PPTOD$_{\text{base}}$+ME} significantly boosts the JGA by 3.1\% on MultiWOZ 2.0 and 2.3\% on MultiWOZ 2.1. \texttt{KAGE-GPT2+ME} also improves its baselines by around 0.9\% on both datasets. Compared with the dialogue-level AL strategy \textbf{CUDS}, our turn-level methods improve the JGA by a large margin (2.3\%$\sim$4.3\% on both datasets).
Considering that DST is a difficult task \citep{budzianowski-etal-2018-multiwoz, wu-etal-2019-transferable, lee-etal-2021-dialogue}, such JGA improvements demonstrate the effectiveness of our turn-level AL framework, which can effectively find the turns that the base DST model can learn the most from. 

\paragraph{Our \textit{turn-level} AL strategy reduces annotation cost.} The reading costs (RC) of \texttt{PPTOD$_{\text{base}}$+ME} and \texttt{KAGE-GPT2+ME} drop by a large margin (around 29\%$\sim$43\%) compared to the Last Turn and CUDS settings, indicating the benefits and necessity of selecting dialogue turns. This significantly saves the annotation cost because a human annotator does not need to read the entire dialogue to label the last turn but only needs to read until the selected turn.

\paragraph{Our approach uses less annotated data can achieve the same or better DST performance.}  To further explore the capability of our AL approach, we plot the intermediate DST performance during the four iterations, as shown in Fig.\ref{k2000_4plot}. Notably, PPTOD$_{\text{base}}$ with Least Confidence (LC) and Maximum Entropy (ME) turn selection methods surpass the Last Turn baselines at just the second or third iteration on MultiWOZ 2.0 and MultiWOZ 2.1 respectively, showing the large data efficiency of our approach (only 7.3\% / 10.9\% data are used). This can be explained that PPTOD$_{\text{base}}$ is fine-tuned on so-far selected turns after each iteration and gains a more robust perception of unseen data, thus tending to choose the turns that are more beneficial to the model. In contrast, KAGE-GPT2 underperforms the Last Turn setting in early iterations, achieving slightly higher accuracy in the final round. Despite this, the overall performance of KAGE-GPT2 is still better than PPTOD$_{\text{base}}$ under the weakly-supervised settings. This is possibly because the additional graph component in KAGE-GPT2 enhances the predictions at intermediate turns and the correlated domain-slots \citep{lin-etal-2021-knowledge}. However, when using CUDS, both DST models underperform a lot on both datasets, especially during early iterations. This indicates that the dialogue-level strategy, which does not distinguish the importance of turns in a dialogue, might not be optimal for selecting training data. In Section \ref{ablation_study}, we show that a smaller query size $k$ can achieve higher data efficiency.

\subsection{Ablation Studies}
\label{ablation_study}

In this section, we further investigate the factors that impact our turn-level AL framework.

% \subsubsection{Effect of Dialogue Query Size}
% \label{ablation_query_size}

\paragraph{Effect of Dialogue Query Size.} Theoretically, the smaller size of queried data per AL iteration, 
the more intermediate models are trained, resulting the better model performance. Moreover, smaller query size is more realistic since the annotation budget is generally limited and there lack enough annotators to label large amount of dialogues after each iteration. To this end, we initialize the unlabelled pool $\mathcal{U}$ by randomly sampling 3,000 dialogues from the MultiWOZ 2.0 training set, and apply our AL framework to KAGE-GPT2, using different query sizes, i.e., $k=500, 1000, 1500$, which leads to $6, 3, 2$ rounds respectively.  

From Fig.\ref{ablation_k_size}, we first observe that smaller $k$ improves the intermediate DST performance: when $k=500$, both LC and ME strategies boost the accuracy by a large margin at the second iteration than $k=1000$, and at the third iteration than $k=1500$. This suggests that, with the same number of training data, the multiple-trained DST model gains the ability to have a more accurate perception of the unseen data. By calculating the prediction uncertainty of the new data, the model tends to choose the turns that it can learn the most from. In contrast, RS chooses a random turn regardless of how many AL rounds, therefore does not show the same pattern as LC and ME. Finally, we find a smaller $k$ tends to achieve higher data efficiency when using LC and ME strategies. It is clear from the figure that $k=500$ uses the least data when reaching the same level of accuracy. However, the drawback of a smaller query size is that it increases overall computation time as more intermediate models have to be trained. 
We provide a computational cost analysis in Section \ref{cost_analysis}.

% \subsubsection{Effect of Base DST Model}
% \label{ablation_base_DST}

\begin{figure}
  \centering
  \includegraphics[width=0.9\columnwidth]{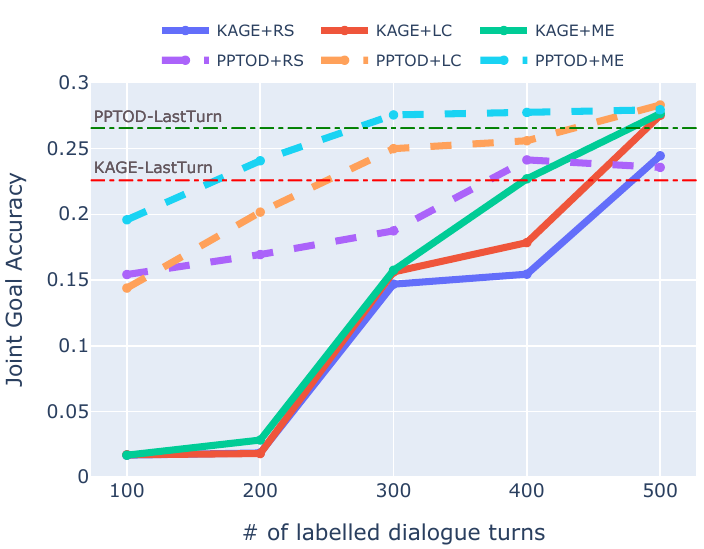}
  \caption{Joint goal accuracy on test sets of KAGE-GPT2 and PPTOD$_{\text{base}}$ on MultiWOZ 2.0 with $k=100$. Results are averaged over three runs.}
  \label{ablation_base_DST_k100}
\end{figure}

\paragraph{Effect of Base DST Model.} It is no doubt that the base DST model is critical to our turn-level AL framework as it directly determines the upper and lower limit of the overall performance. However, we are interested to see how our approach can further boost the performance of different DST models. We randomly sample $\mathcal{U}=500$ dialogues from the MultiWOZ 2.0 training set and set the query size $k=100$ for both models. As shown in Fig.\ref{ablation_base_DST_k100}, we also report the results of the two models using the non-AL strategy of Last Turn, which can be considered as the lower performance baselines.

% We first notice that PPTOD$_{\text{base}}$ outperforms KAGE-GPT2 in the Last Turn setting when using only 500 training data, showing the superiority of PPTOD$_{\text{base}}$ under the extreme low-resource scenario. 

We first confirm that both PPTOD$_{\text{base}}$ and KAGE-GPT2 outperform their Last Turn baselines after applying our AL framework, demonstrating both data efficiency and effectiveness of our approach. Secondly, we notice that PPTOD$_{\text{base}}$ achieves comparable accuracy in the first two rounds, while KAGE-GPT2 nearly stays at 0 regardless of the turn selection methods, showing the superiority of PPTOD$_{\text{base}}$ under the extreme low-resource scenario. This is possibly because PPTOD$_{\text{base}}$ is pre-trained on large dialogue corpora thus gains few-shot learning ability \citep{su-etal-2022-multi}, whereas only 200 training data are not enough for KAGE-GPT2 to be fine-tuned. However, in the later iterations, the performance of KAGE-GPT2 grows significantly, especially when using the ME strategy, eventually reaching the same level as PPTOD$_{\text{base}}$. In contrast, the accuracy of PPTOD$_{\text{base}}$ increases slowly, indicating the model gradually becomes insensitive to the newly labelled data.

% In contrast, the accuracy of PPTOD$_{\text{base}}$ increases slowly, indicating the model is insensitive to unseen data.

% With only 400 instances, \texttt{KAGE-GPT2+ME} surpasses its baseline and reaches the same level of PPTOD$_{\text{base}}$ in the final round, 

% \subsubsection{Effect of Turn Selection Strategy}
% \label{ablation_turn_selection_method}

\begin{table}
\centering
% \addtolength{\tabcolsep}{-4.5pt}
\resizebox{0.7\columnwidth}{!}{
% \small
\begin{tabular}{c|cc}
\hline
\textbf{Method} & KAGE-GPT2  & PPTOD$_{\text{base}}$ \\
\hline
% RS & 57.83$\pm28.1$ & 57.92$\pm30.4$ \\
LC & 76.51\tiny{$\pm24.7$} & 81.13\tiny{$\pm22.3$} \\
ME & 68.18\tiny{$\pm29.1$} & 58.68\tiny{$\pm31.5$}  \\ 
\hline
\end{tabular}
}
\caption{Reading Cost (RC) (\%) of different turn selection methods. The lower the better.}
\label{k100_reading_cost}
\end{table}

% From Figure \ref{k2000_4plot}, while ME and LC improve over the RS baseline, they do not consistently outperform each other \textcolor{red}{Ling: which one? or the others?} during AL iterations in terms of the joint goal accuracy. However, as shown in Table \ref{main_results}, LC results in a higher Reading Cost (RC) than ME, which means LC tends to select latter half of turns in dialogues. A visualization of the distributions of RC in Appendix \ref{appendix_visualization} also confirms this finding. Moreover, ME achieves the best results in all final rounds. From Figure \ref{ablation_base_DST_k100}, we find that ME is consistently better than LC for both DST models, which demonstrates the effectiveness of ME under small query size $k$. We report their RC in Table \ref{k100_reading_cost}, which also confirms that ME saves reading costs than LC. We present some examples of the turns selected by ME and LC in Appendix \ref{appendix_examples}.

%%%%%%%% Example 1 %%%%%%%%
\begin{table*}[ht]
% \small
\caption{Example (MUL0295) of the selected turn (marks by \checkmark) by PPTOD$_{\text{base}}$ using ME and LC.}
\centering
\resizebox{0.9\textwidth}{!}{
\begin{tabular}{clcclll}
\cline{1-4}
\multicolumn{1}{l}{} & \multicolumn{1}{c|}{\textbf{Dialogue MUL0295}} & \multicolumn{1}{c|}{\textbf{ME}} & \textbf{LC} &  &  &  \\ \cline{1-4}
Turn 1 & \multicolumn{1}{l|}{\begin{tabular}[c]{@{}l@{}}{[}S{]}: \\ {[}U{]}: i am looking for an expensive place to dine in the centre of town.\\ \textit{State: \{restaurant-area=centre, restaurant-pricerange=expensive\}}\end{tabular}} & \multicolumn{1}{c|}{} &  &  &  &  \\ \cline{1-4}
Turn 2 & \multicolumn{1}{l|}{\begin{tabular}[c]{@{}l@{}}{[}S{]}: great kymmoy is in the centre of town and expensive.\\ {[}U{]}: i want to book a table for 3 people at 14:00 on Saturday.\\ \textit{State: \{restaurant-book day=saturday, restaurant-book people=3, restaurant-book time=14:00\}}\end{tabular}} & \multicolumn{1}{c|}{} &  &  &  &  \\ \cline{1-4}
Turn 3 & \multicolumn{1}{l|}{\begin{tabular}[c]{@{}l@{}}{[}S{]}: booking was successful. the table will be reserved for 15 minutes. reference number is: vbpwad3j.\\ {[}U{]}: thank you so much. i would also like to find a train to take me to kings lynn by 10:15.\\ \textit{State: \{train-destination=kings lynn, train-arriveby=10:15\}}\end{tabular}} & \multicolumn{1}{c|}{} & \checkmark &  &  &  \\ \cline{1-4}
Turn 4 & \multicolumn{1}{l|}{\begin{tabular}[c]{@{}l@{}}{[}S{]}: there are 35 departures with those criteria. what time do you want to leave?\\ {[}U{]}: the train should arrive by 10:15 please on sunday please.\\ \textit{State: \{train-day=sunday\}}\end{tabular}} & \multicolumn{1}{c|}{\checkmark} &  &  &  &  \\ \cline{1-4}
Turn 5 & \multicolumn{1}{l|}{\begin{tabular}[c]{@{}l@{}}{[}S{]}: how many tickets will you need?\\ {[}U{]}: just 1 ticket. i will need the train id, cost of ticket and exact departure time as well.\\ \textit{State: \{\}}\end{tabular}} & \multicolumn{1}{c|}{} &  &  &  &  \\ \cline{1-4}
Turn 6 & \multicolumn{1}{l|}{\begin{tabular}[c]{@{}l@{}}{[}S{]}: there is a train arriving in kings lynn on sunday at 09:58. it departs at 09:11 and costs 7.84 pounds. the train id is tr6088.\\ {[}U{]}: great! that s all i needed. thanks a lot for the help.\\ \textit{State: \{\}}\end{tabular}} & \multicolumn{1}{c|}{} &  &  &  &  \\ 
\cline{1-4}
\end{tabular}
}
\label{example_1}
\end{table*}

% ME does not consistenly outperform LC, and vice visa

\begin{figure}[ht]
  \centering
  \includegraphics[width=0.8\columnwidth]{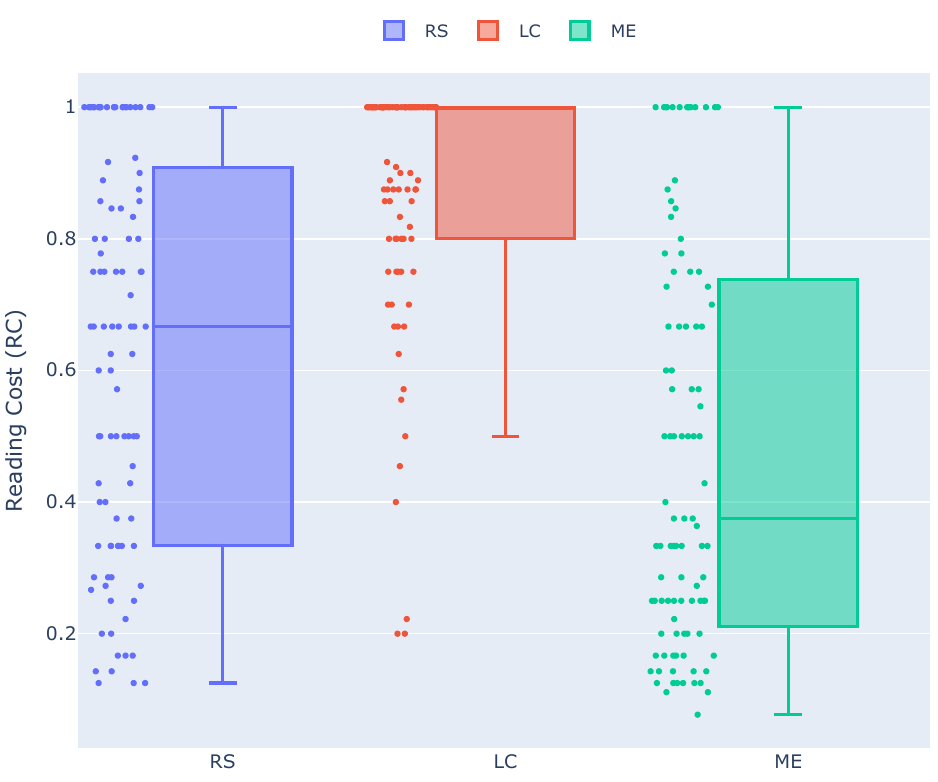}
  \caption{Visualization of the turns selected by PPTOD$_{\text{base}}$ at the final round ($k=100$). ME reduces RC the most.
  }
  \label{visualization_pptod_body}
\end{figure}

% compare runtime
\begin{table}
\centering
% \addtolength{\tabcolsep}{-4.5pt}
\resizebox{\columnwidth}{!}{
% \small
\begin{tabular}{c|cccc}
\hline
\textbf{Method} & \textbf{\# of Training data (\%)} $\downarrow$ & \textbf{JGA} $\uparrow$ & \textbf{RC} $\downarrow$ & \textbf{Runtime (hour)} $\downarrow$\\
\hline
Full data & 21072 (100\%)                & 46.7    & 100    & 2.3                     \\
Last Turn & 3000 (14.2\%)                & 41.4    & 100    & 0.6                     \\
ME  & 3000 (14.2\%)                & 44.3    & 59.3   & 1.6 \\ 
\hline
\end{tabular}
}
\caption{Computational cost comparison using KAGE-GPT2 on MultiWOZ 2.0 with $\mathcal{U} = 3000$ and $k=1000$.}
\label{table_runtime_cost}
\end{table}

\begin{table}
\centering
% \addtolength{\tabcolsep}{-4.5pt}
\resizebox{\columnwidth}{!}{
% \small
\begin{tabular}{c|cc}
\hline
\textbf{Method} & \textbf{Total Annotation Cost (\$)} $\downarrow$  \\
\hline
Full Dialogue & $z*(T*x + T*y)$ \\
Last Turn & $z*(T*x + 1*y)$  \\
Selected Turn (Ours) & $z*(t*x + 1*y)$, where $ 1 \leq t \leq T$ \\ 
\hline
\end{tabular}
}
\caption{Annotation cost estimation comparison of different methods. 
% $x$ represents reading time (minutes) for a human annotator to read each turn, $y$ represents annotating time (minutes) to label a single turn, $z$ represents the cost of hiring a human annotator (\$/minute), $T$ indicates the total number of turns in a dialogue.
}
\label{table_computation_cost}
\end{table}
 
\paragraph{Effect of Turn Selection Strategy.} From Fig.\ref{k2000_4plot}, while both ME and LC improve over the RS baseline, ME does not consistently outperform LC during AL iterations in terms of the joint goal accuracy, and vice versa. 
However, as shown in Table \ref{main_results}, LC results in a higher Reading Cost (RC) than ME, which means LC tends to select latter half of turns in dialogues. 
Conversely, ME significantly reduces RC in the last iteration (Fig.\ref{visualization_pptod_body}; more in Appendix \ref{appendix_visualization}) and is consistently better than LC and RS for both DST models (Fig.\ref{ablation_base_DST_k100}), which demonstrates the effectiveness of ME under small query size $k$.  
% A visualization of the distributions of RC in Appendix \ref{appendix_visualization} also confirms this finding. 
% Moreover, ME achieves the best results in all final rounds. From Figure \ref{ablation_base_DST_k100}, we find that ME is consistently better than LC for both DST models, which demonstrates the effectiveness of ME under small query size $k$. 
We report their RC in Table \ref{k100_reading_cost}, which also confirms that ME saves reading costs than LC. 
An example of the turns selected by ME and LC in a dialogue is shown in Table \ref{example_1}, more examples in Appendix \ref{appendix_examples}.

\subsection{Cost Analysis}
\label{cost_analysis}

Our AL-based method saves annotation costs and achieves comparable DST performance with traditional methods at the expense of increased computation time.
In this section, we conduct a cost analysis, including computation and annotation costs. 
We initialize the unlabelled pool $\mathcal{U}$ by randomly sampling 3,000 dialogues from the MultiWOZ 2.0 training set, and apply our AL framework to KAGE-GPT2, and set the query size as $k=1000$. 
As shown in Table \ref{table_runtime_cost}, our method improves JGA and RC than the Last Turn baseline, but with an increased runtime since our method requires three rounds of iteration.

% In practice, the number of annotators depends on the financial budget, project timeline, and the proficiency of annotators. Estimating the exact number of annotators and the annotation cost is challenging.
Due to a lack of budget, we are unable to employ human annotators to evaluate the actual annotation cost. Instead, we conduct a theoretical cost analysis to show the potential cost reduction of our method.
Suppose a dialogue $D$ has $T$ turns in total, and it takes $x$ minutes for a human annotator to read each turn (\textit{i.e.}, reading time), $y$ minutes to annotate a single turn (\textit{i.e.}, annotating time), $z$ dollars per minute to hire a human annotator.
Assuming our proposed method selects the $t$th ($ 1 \leq t \leq T$) turn to annotate.
The total annotation cost, including the reading time and annotating time of three methods, are listed in Table \ref{table_computation_cost}.
Since the Full Dialogue baseline takes each accumulated turn as a training instance (Section \ref{preliminaries}), it requires the highest annotation cost. 
Our method only annotates a single turn per dialogue, the same as the Last Turn baseline. Therefore, the annotation cost lies in the selected turn $t$, which is measured by RC in our experiments. 
As shown in Table \ref{main_results} and discussed in Section \ref{sec_main_results}, our method generally saves RC by a large margin (around 29\%$\sim$43\% across different models) compared to the Last Turn baseline and saves more compared to the Full data setting.
Therefore, from a theoretical cost estimation point of view, our proposed method can save annotation costs while maintaining DST performance.

\section{Conclusion}

% This paper tackles the practical dialogue annotation problem by proposing a novel turn-level AL framework for DST, which strategically selects the most valuable turn from each dialogue for labelling and training. Experiments show that our approach outperforms strong DST baselines in the weakly-supervised scenarios and achieves the same or better joint goal and slot accuracy \textcolor{red}{Ling: how about slot accuracy?} with significantly less annotated data. Further analysis demonstrates that our approach is not only effective but also provides a cost-saving way to annotate new dialogue data.\textcolor{red}{Ling: not sure what you mean...} In the future, we are interested in exploring an AL query strategy specific to DST and experimenting with the effect of the warm-start setting. We will also study on the combination of dialogue-level and turn-level AL for DST. \textcolor{red}{Ling: how about the combination of dialogue level and turn level AL?}

This paper tackles the practical dialogue annotation problem by proposing a novel turn-level AL framework for DST, which strategically selects the most valuable turn from each dialogue for labelling and training. Experiments show that our approach outperforms strong DST baselines in the weakly-supervised scenarios and achieves the same or better joint goal and slot accuracy with significantly less annotated data. Further analysis are conducted to investigate the impact of AL query sizes, base DST models and turn selection methods. 
% In the future, we are interested in combining the dialogue-level and turn-level AL query strategy.

% exploring an AL query strategy specific to DST by 
%In the future, we are interested in exploring an AL query strategy specific to DST, combining dialogue-level and turn-level selection, and experimenting with the effect of the warm-start setting. 

%  We will also study on the combination of dialogue-level and turn-level AL for DST.

\section{Limitations}
\label{limitations}

We acknowledge the limitations of this paper as follows.

First, our AL approach adds extra computation time compared to directly training a DST model using only the last turns of dialogues. A smaller query size (e.g., $k$) may further increase the runtime as more intermediate models have to be trained. That is, we achieved similar or even better DST performance with significantly reduced annotation data at the cost of increased computation time. Therefore, the trade-off between computational cost, DST performance, and annotation cost needs to be well-determined. 

Second, we are unable to employ human annotators to evaluate the actual cost due to a lack of budget.
In practice, the number of annotators required depends on the financial budget, project timeline, and the proficiency of annotators. Estimating the exact number of annotators and the annotation cost is challenging.
As a mitigation, we provide a theoretical cost analysis in Section \ref{cost_analysis}. However, it is a rough estimation and may not reflect the actual cost.

Third, our experiments are limited to the MultiWOZ 2.0 \citep{budzianowski-etal-2018-multiwoz} and MultiWOZ 2.1 \citep{eric-etal-2020-multiwoz} datasets.
We also tried to use the SGD dataset \citep{Rastogi_Zang_Sunkara_Gupta_Khaitan_2020}. However, the PPTOD model is already pre-trained on this dataset, making it unsuitable for downstream evaluation. KAGE-GPT2 requires the predefined ontology (i.e., the all possible domain-slot value pairs in the dataset) to build a graph neural network, but SGD does not provide all possible values for non-categorical slots.
For example, MultiWOZ has all possible values predefined for the non-categorical domain-slot \textit{train-arriveBy}, while SGD does not have it since it is innumerable.
Our AL framework is built upon the base DST model and thus suffers the same drawbacks; we may try other DST models and datasets in the future.

\section*{Acknowledgements}

This work is supported by TPG Telecom. 
We would like to thank anonymous reviewers for their valuable comments. 

\bibliography{anthology,custom}
\bibliographystyle{acl_natbib}

\clearpage
\onecolumn
\appendix

\section{Datasets Statistics}
\label{appendix_datasets}

\begin{table}[H]\centering
\resizebox{0.7\columnwidth}{!}{
\begin{tabular}{clcc}
\hline
\multicolumn{1}{l}{}                             &                            & MultiWOZ2.0 & MultiWOZ2.1 \\ \hline
\multicolumn{1}{c|}{\multirow{8}{*}{Train}}      & \# Dialogues               & 7888        & 7888        \\
\multicolumn{1}{c|}{}                            & \# Domains                 & 5           & 5           \\
\multicolumn{1}{c|}{}                            & \# Slots                   & 30          & 30          \\
\multicolumn{1}{c|}{}                            & \# Total turns             & 54945       & 54961       \\
\multicolumn{1}{c|}{}                            & \# Last turns              & 7888        & 7888        \\
\multicolumn{1}{c|}{}                            & \# Avg. turns per dialogue & 6.97        & 6.97        \\
\multicolumn{1}{c|}{}                            & \# Max turns per dialogue  & 22          & 22           \\
\multicolumn{1}{c|}{}                            & \# Min turns per dialogue  & 1           &  1         \\ \hline
\multicolumn{1}{c|}{\multirow{2}{*}{Validation}} & \# Dialogues               & 1000        & 1000        \\
\multicolumn{1}{c|}{}                            & \# Total turns             & 7374        & 7374        \\ \hline
\multicolumn{1}{c|}{\multirow{2}{*}{Test}}       & \# Dialogues               & 1000        & 999         \\
\multicolumn{1}{c|}{}                            & \# Total turns             & 7372        & 7368        \\ \hline
\end{tabular}
}
\caption{\label{datasets_statistics}
% \small
Statistics of the datasets in the experiments.
}
\end{table}

\section{Configuration Details}
\label{appendix_config_details}

We use the official release of KAGE-GPT2\footnote{\url{https://github.com/LinWeizheDragon/Knowledge-Aware-Graph-Enhanced-GPT-2-for-Dialogue-State-Tracking}} \citep{lin-etal-2021-knowledge} and PPTOD\footnote{\url{https://github.com/awslabs/pptod}} \citep{su-etal-2022-multi} to implement our turn-level AL framework. 
% All experiments were done with a NVIDIA T4 GPU.

\paragraph{KAGE-GPT2} We use the \texttt{L4P4K2-DSGraph} model setup and follow its sparse supervision (last turn) hyperparameter settings. Specifically, the loaded pre-trained GPT-2 model has 12 layers, 768 hidden size, 12 heads and 117M parameters, which is provided by HuggingFace\footnote{\url{https://huggingface.co/models}}. AdamW optimizer with a linear decay rate $1 \times 10^{-12}$ is used when training. The GPT-2 component and the graph component are jointly trained, with the initial learning rates are $6.25 \times 10^{-5}$ and $8 \times 10^{-5}$ respectively. The training batch size used is 2, while the batch size for validation and evaluation is 16.

\paragraph{PPTOD} We use the released \texttt{base} checkpoint, which is initialized with a T5-base model with around 220M parameters. PPTOD$_{\text{base}}$ is pre-trained on large dialogue corpora, for more details, we refer readers to the original paper. When training, Adafactor optimizer is used and the learning rate is $1 \times 10^{-3}$. Both training, validation, and evaluation batch size used is 4.

\paragraph{Turn Selection} During each AL iteration, we use the trained model from the last iteration to evaluate all the turns within a dialogue and then select a turn based on the acquisition strategy. 

\paragraph{Training} At the end of each iteration, we re-initialize a new pre-trained GPT-2 model for KAGE-GPT2 or re-initialize a new model from the released pre-trained \texttt{base} checkpoint for PPTOD, and then train the model as usual with all current accumulated labelled turns. We train the DST model for 150 epochs using the current accumulated labelled pool $\mathcal{L}$, and early stop when the performance is not improved for 5 epochs on the validation set. Importantly, instead of using the full 7,374 validation set, we only use the last turn of each dialogue to simulate the real-world scenario, where a large amount of annotated validation set is also difficult to obtain \citep{perez2021true}. However, we use the full test set when evaluating.

% During each AL iteration, we use the trained model from the last iteration to evaluate all the turns within a dialogue and then select a turn based on the acquisition strategy. At the end of each iteration, we re-initialize a new model for KAGE-GPT2 or re-initialize a new model from the released pre-trained \texttt{base} checkpoint for PPTOD, and then train the model as usual with all current accumulated labelled turns. 

% \section{Additional Results}
% \label{appendix_additional}

\section{Visualization of Selected Turns}
\label{appendix_visualization}

% \begin{table}
% \centering
% % \addtolength{\tabcolsep}{-4.5pt}
% \resizebox{0.4\columnwidth}{!}{
% \small
% \begin{tabular}{c|cc}
% \hline
% \textbf{Method} & KAGE-GPT2  & PPTOD$_{\text{base}}$ \\
% \hline
% % RS & 57.83$\pm28.1$ & 57.92$\pm30.4$ \\
% LC & 76.51\tiny{$\pm24.7$} & 81.13\tiny{$\pm22.3$} \\
% ME & 68.18\tiny{$\pm29.1$} & 58.68\tiny{$\pm31.5$}  \\ 
% \hline
% \end{tabular}
% }
% \caption{Reading Cost (RC) (\%) of different turn selection methods. The lower the better.}
% \label{k100_reading_cost}
% \end{table}

\begin{figure}[ht]
  \centering
  \includegraphics[width=0.5\columnwidth]{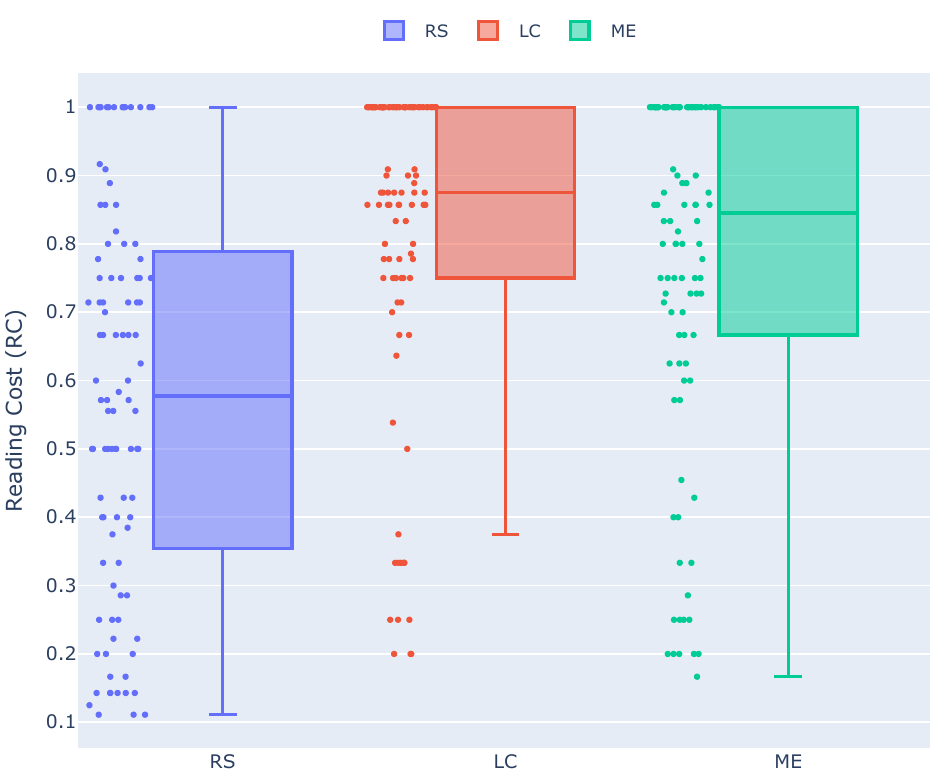}
  \caption{Visualization of the turns selected by KAGE-GPT2 at the final round ($k=100$).
  }
  \label{visualization_kage}
\end{figure}

To clearly compare the reading costs of different turn selection methods, we visualize the distributions of the selected turns at the final round for the setting in Section \ref{ablation_study}, as shown in Fig.\ref{visualization_pptod_body} and Fig.\ref{visualization_kage}. A dot means a selected turn from a dialogue, while the ends of the box represent the lower and upper quartiles, and the median (second quartile) is marked by a line inside the box. A higher RC means the turn is selected from the second half of the conversation ($\text{RC}=1$ means the last turn is selected); thus, a human annotator needs to read most of the conversation to label its state, which is more costly. From the figures, overall, RS distributes randomly, while ME has a much lower reading cost than LC, especially for PPTOD$_{\text{base}}$.

% \begin{figure}[ht]
%   \centering
%   \includegraphics[width=0.5\columnwidth]{visualization_pptod.pdf}
%   \caption{Visualization of the turns selected by PPTOD$_{\text{base}}$ at the final round ($k=100$).
%   }
%   \label{visualization_pptod}
% \end{figure}

\section{Example of Selected Turns}
\label{appendix_examples}

Table \ref{example_1}, Table \ref{example_2} and Table \ref{example_3} present the examples of selected turns by ME and LC using PPTOD$_{\text{base}}$ from MultiWOZ 2.0. \text{[S]} and \text{[U]} denote the system and user utterance respectively, while \textit{State} represents the dialogue states that are mentioned at the current turn. \checkmark marks the selected turn by the strategy and is the only turn in the dialogue used for training. Although not always the case, we can see that both ME and LC can select the earliest turn that summarizes the entire dialogue, which not only saves the need to read through the whole conversation but also keeps the valuable context information intact as much as possible. However, still, a more suitable AL query strategy for DST is worthy of being studied.

\begin{table}[ht]
% \small
\caption{Example (MUL1068) of the selected turn by PPTOD$_{\text{base}}$ using ME and LC.}
\centering
\resizebox{1.0\textwidth}{!}{
\begin{tabular}{clcclll}
\cline{1-4}
\multicolumn{1}{l}{} & \multicolumn{1}{c|}{\textbf{Dialogue MUL1068}} & \multicolumn{1}{c|}{\textbf{ME}} & \textbf{LC} &  &  &  \\ \cline{1-4}
Turn 1 & \multicolumn{1}{l|}{\begin{tabular}[c]{@{}l@{}}{[}S{]}: \\ {[}U{]}: i am looking for a gueshouse to stay in during my visit. i would like for it to be in the east section of town.\\ \textit{State: \{hotel-area=east, hotel-type=guesthouse\}}\end{tabular}} & \multicolumn{1}{c|}{} &  &  &  &  \\ \cline{1-4}
Turn 2 & \multicolumn{1}{l|}{\begin{tabular}[c]{@{}l@{}}{[}S{]}: there are 6 entries, do you have any other preferences in hotel choice?\\ {[}U{]}: i would like a 4 star place that's on the pricier side.\\ \textit{State: \{hotel-pricerange=expensive, hotel-stars=4\}}\end{tabular}} & \multicolumn{1}{c|}{} &  &  &  &  \\ \cline{1-4}
Turn 3 & \multicolumn{1}{l|}{\begin{tabular}[c]{@{}l@{}}{[}S{]}: there are no 4 star expensive guesthouses on the east side of town. would you be interested in a cheap or moderate -ly priced 1 instead?\\ {[}U{]}: i would like 1 that is cheap then.\\ \textit{State: \{hotel-pricerangen=cheap\}}\end{tabular}} & \multicolumn{1}{c|}{} & &  &  &  \\ \cline{1-4}
Turn 4 & \multicolumn{1}{l|}{\begin{tabular}[c]{@{}l@{}}{[}S{]}: autumn house is cheap and has free internet and parking . would you like me to book a reservation for you?\\ {[}U{]}: yes please. can you book the reservation for saturday for 5 guests for a 2 night stay. thanks.\\ \textit{State: \{hotel-book day=saturday, hotel-book people=5, hotel-book stay=2\}}\end{tabular}} & \multicolumn{1}{c|}{} &  &  &  &  \\ \cline{1-4}
Turn 5 & \multicolumn{1}{l|}{\begin{tabular}[c]{@{}l@{}}{[}S{]}: sorry, there are no rooms available for your specification -s. do you want to try a different day or a shorter stay?\\ {[}U{]}: try for 1 night instead. if that works, can i have the reference number?\\ \textit{State: \{hotel-book stay=1\}}\end{tabular}} & \multicolumn{1}{c|}{} &  &  &  &  \\ \cline{1-4}
Turn 6 & \multicolumn{1}{l|}{\begin{tabular}[c]{@{}l@{}}{[}S{]}: how about the autumn house for 1 night starting saturday and your reference number is svciwlwa.\\ {[}U{]}: i also need to find something called gonville and caius college, can you help?\\ \textit{State: \{attraction-name=gonville and caius college\}}\end{tabular}} & \multicolumn{1}{c|}{\checkmark} &  &  &  &  \\ \cline{1-4}
Turn 7 & \multicolumn{1}{l|}{\begin{tabular}[c]{@{}l@{}}{[}S{]}: gonville and caius college is located in the center of town on trinity street . anything else i can do for you?\\ {[}U{]}: can i get the postcode ?\\ \textit{State: \{\}}\end{tabular}} & \multicolumn{1}{c|}{} &  &  &  &  \\ \cline{1-4}
Turn 8 & \multicolumn{1}{l|}{\begin{tabular}[c]{@{}l@{}}{[}S{]}: no problem , the postal code is cb21ta. did you need the phone number as well?\\ {[}U{]}: no thanks i am all set . thank you for your help today.\\ \textit{State: \{\}}\end{tabular}} & \multicolumn{1}{c|}{} & \checkmark  &  &  &  \\ \cline{1-4}
% \multicolumn{1}{l}{} &  & \multicolumn{1}{l}{} & \multicolumn{1}{l}{} &  &  &  \\
% \multicolumn{1}{l}{} &  & \multicolumn{1}{l}{} & \multicolumn{1}{l}{} &  &  &  \\
% \multicolumn{1}{l}{} &  & \multicolumn{1}{l}{} & \multicolumn{1}{l}{} &  &  &  \\
% \multicolumn{1}{l}{} &  & \multicolumn{1}{l}{} & \multicolumn{1}{l}{} &  &  & 
\end{tabular}
}
\label{example_2}
\end{table}

%%%%%%%% Example 3 %%%%%%%%
\begin{table}[ht]
% \small
\caption{Example (PMUL2281) of the selected turn by PPTOD$_{\text{base}}$ using ME and LC.}
\centering
\resizebox{1.0\textwidth}{!}{
\begin{tabular}{clcclll}
\cline{1-4}
\multicolumn{1}{l}{} & \multicolumn{1}{c|}{\textbf{Dialogue PMUL2281}} & \multicolumn{1}{c|}{\textbf{ME}} & \textbf{LC} &  &  &  \\ \cline{1-4}
Turn 1 & \multicolumn{1}{l|}{\begin{tabular}[c]{@{}l@{}}{[}S{]}: \\ {[}U{]}: can you help me find a place to dine?\\ \textit{State: \{\}}\end{tabular}} & \multicolumn{1}{c|}{} &  &  &  &  \\ \cline{1-4}
Turn 2 & \multicolumn{1}{l|}{\begin{tabular}[c]{@{}l@{}}{[}S{]}: sure! what type of food would you like to eat in what area?\\ {[}U{]}: i would like some north indian food that is expensive, in the south.\\ \textit{State: \{restaurant-food=north indian\}}\end{tabular}} & \multicolumn{1}{c|}{} &  &  &  &  \\ \cline{1-4}
Turn 3 & \multicolumn{1}{l|}{\begin{tabular}[c]{@{}l@{}}{[}S{]}: we do not have any north indian restaurant -s, though we do have many indian restaurant -s.\\ {[}U{]}: indian food would be fine then, can you tell me the name?\\ \textit{State: \{restaurant-pricerange=expensive, restaurant-food=indian\}}\end{tabular}} & \multicolumn{1}{c|}{} & &  &  &  \\ \cline{1-4}
Turn 4 & \multicolumn{1}{l|}{\begin{tabular}[c]{@{}l@{}}{[}S{]}: there are several indian restaurant -s, may i suggest the golden curry in the centre. it sounds like just what you are looking for.\\ {[}U{]}: i was actually hoping for a restaurant in the south. are there any available?\\ \textit{State: \{restaurant-area=south\}}\end{tabular}} & \multicolumn{1}{c|}{} & \checkmark &  &  &  \\ \cline{1-4}
Turn 5 & \multicolumn{1}{l|}{\begin{tabular}[c]{@{}l@{}}{[}S{]}: i found 1 called taj tandoori. want more information?\\ {[}U{]}: more information please\\ \textit{State: \{\}}\end{tabular}} & \multicolumn{1}{c|}{} &  &  &  &  \\ \cline{1-4}
Turn 6 & \multicolumn{1}{l|}{\begin{tabular}[c]{@{}l@{}}{[}S{]}: it is expensive and in the south.\\ {[}U{]}: can i get the postcode for that please?\\ \textit{State: \{\}}\end{tabular}} & \multicolumn{1}{c|}{\checkmark} &  &  &  &  \\ \cline{1-4}
Turn 7 & \multicolumn{1}{l|}{\begin{tabular}[c]{@{}l@{}}{[}S{]}: sure! the post code is cb17aa .\\ {[}U{]}: thanks! i am also looking for a nightclub. i'll need the postcode, please?\\ \textit{State: \{attraction-type=nightclub\}}\end{tabular}} & \multicolumn{1}{c|}{} &  &  &  &  \\ \cline{1-4}
Turn 8 & \multicolumn{1}{l|}{\begin{tabular}[c]{@{}l@{}}{[}S{]}: the ballare is a nightclub in the centre of town. the entrance fee is 5 pounds.\\ {[}U{]}: i'll try that. what s the postcode please?\\ \textit{State: \{\}}\end{tabular}} & \multicolumn{1}{c|}{} &  &  &  &  \\ \cline{1-4}
Turn 9 & \multicolumn{1}{l|}{\begin{tabular}[c]{@{}l@{}}{[}S{]}: their postcode is cb23na. can i help you with anything else today?\\ {[}U{]}: no thanks. that was all i needed today. goodbye.\\ \textit{State: \{\}}\end{tabular}} & \multicolumn{1}{c|}{} &  &  &  &  \\ \cline{1-4}
% \multicolumn{1}{l}{} &  & \multicolumn{1}{l}{} & \multicolumn{1}{l}{} &  &  &  \\
% \multicolumn{1}{l}{} &  & \multicolumn{1}{l}{} & \multicolumn{1}{l}{} &  &  &  \\
% \multicolumn{1}{l}{} &  & \multicolumn{1}{l}{} & \multicolumn{1}{l}{} &  &  &  \\
% \multicolumn{1}{l}{} &  & \multicolumn{1}{l}{} & \multicolumn{1}{l}{} &  &  & 
\end{tabular}
}
\label{example_3}
\end{table}

\end{document}